\documentclass[]{spie}  

\usepackage{amsmath,amsfonts,amssymb}
\usepackage{graphicx}
\usepackage[colorlinks=true, allcolors=blue]{hyperref}
\usepackage{svg}

\title{Text controllable PET denoising}

\author[a]{Xuehua Ye}
\author[b]{Hongxu Yang}
\author[c]{Adam J. Schwarz}
\affil[a]{GE Healthcare, China}
\affil[b]{GE Healthcare, Netherlands}
\affil[c]{GE Healthcare, USA}
\authorinfo{Corresponding author: Xuehua Ye, Hongxu Yang (\{Xuehua.Ye, Hongxu.Yang\}@gehealthcare.com)}

\begin{document} 
\maketitle

\begin{abstract}
Positron Emission Tomography (PET) imaging is a vital tool in medical diagnostics, offering detailed insights into molecular processes within the human body. However, PET images often suffer from complicated noise, which can obscure critical diagnostic information. The quality of the PET image is impacted by various factors including scanner hardware, image reconstruction, tracer properties, dose/count level, and acquisition time. In this study, we propose a novel text-guided denoising method capable of enhancing PET images across a wide range of count levels within a single model. The model utilized the features from a pretrained CLIP model with a U-Net based denoising model. Experimental results demonstrate that the proposed model leads significant improvements in both qualitative and quantitative assessments. The flexibility of the model shows the potential for helping more complicated denoising demands or reducing the acquisition time. 
\end{abstract}


\section{INTRODUCTION}

Positron Emission Tomography (PET) is a vital imaging tool for visualizing metabolic and molecular activity. However, low radiotracer doses or short acquisition times, commonly in pediatric or oncology patients, can lead to noisy images and reduced diagnostic accuracy. To mitigate this, PET denoising techniques aim to reconstruct high-quality, full-dose-equivalent images from low-dose (more precisely, low-count) scans~\cite{xie2023ddpet3d}. Traditional methods like Gaussian smoothing often blur fine details, while deep learning models better enhance image quality and preserve anatomical accuracy~\cite{jiang2023petdiffusion, pan2024petconsistency}. Given the rapid growth of large language models in recent years, a new direction involves integrating language and vision-language models into denoising. Recent studies using text to guide to denoise different organs or to reconstruct full dose from two different low-count levels PET images~\cite{yu2025textguided, yu2025petdenoising}. Building on this foundation, we propose a novel text-controlled PET denoising framework that uses semantic cues from count-level descriptions, enabling flexible and controllable denoising output beyond fixed-pair training.

\section{METHODS}
The proposed method integrates textual count information into the image denoising process, enabling it to adaptively handle varying count levels in PET scans. To incorporate semantic understanding of count levels, we leverage the pre-trained CLIP model, which has been trained on a large corpus of image-text pairs~\cite{radford2021clip}. The CLIP text encoder is used to encode textual descriptions of count levels (e.g., “a 1/100 count level PET image”) into high-dimensional embedding vectors, which capture rich semantic information. In architecture as illustrated in Figure 1, the proposed architecture consists of a dual text embedding pathway. For the encoder embedding path, the input count description is encoded using the CLIP text encoder. The resulting embedding is then broadcast and element-wise multiplied with the feature maps at each layer of the U-Net encoder. This operation conditions the encoding process on the specified count level, allowing the model to extract features that are sensitive to the input count context. In the decoder embedding path, a parallel embedding, derived from the output count description, is similarly integrated into each layer of the U-Net decoder. This guides the reconstruction process and is able to help generate output images that align with the desired count level characteristics. By embedding count-level semantics into both the encoding and decoding stages, the model learns to differentiate between input PET images acquired at different count levels and to generate denoised outputs that reflect the characteristics of a target count level (Note that we constrain the output count level to be higher than the input). This design enables flexible and controllable denoising, which is particularly valuable in clinical scenarios where count levels vary due to patient-specific or protocol-specific constraints. 

\begin{figure} [htbp]
\begin{center}
\begin{tabular}{c} 
\includegraphics[width=16cm]{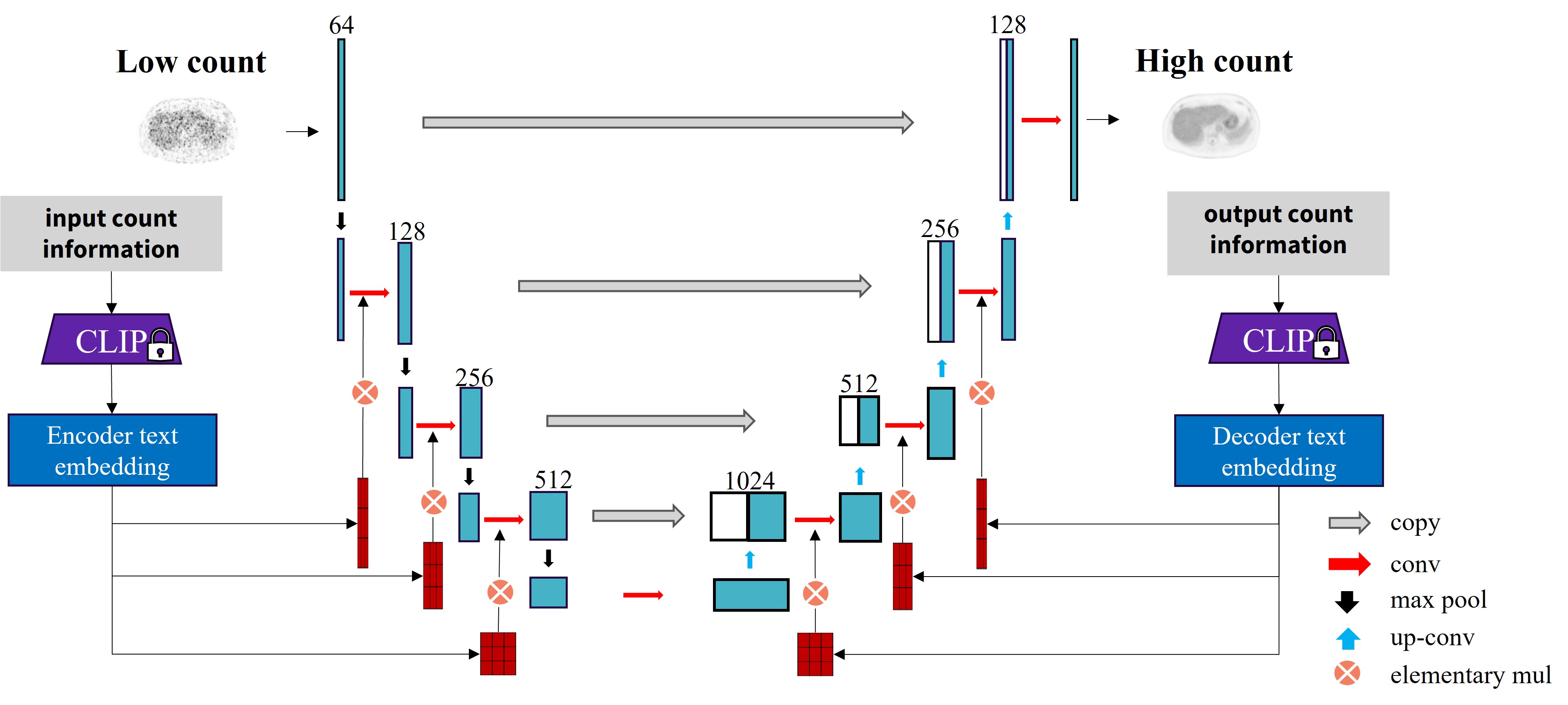}
\end{tabular}
\end{center}
\caption[Fig1]{Proposed text-controlled denoising model. }
\end{figure}

To train the proposed model, the dataset was randomly split into 80\% for training and 20\% for testing, ensuring that patient-level separation was maintained to prevent data leakage. For count-level conditioning, we utilized the CLIP ViT-B/32~\cite{radford2021clip} pretrained model to encode textual descriptions of count levels (e.g., “a 1/100 count level PET image”) into high-dimensional semantic embeddings. During training, each input sample consists of low-count PET axial slice, a corresponding text embedding representing the count level and the higher-count PET axial slice as the ground truth. To promote generalization across count levels, for the input the ground truth, the count condition was randomly sampled at each training step, ensuring the model learns to adapt to a wide range of input and output conditions. The model was trained using two NVIDIA A100 GPUs. We employed the AdamW optimizer with a learning rate of 0.001 and a batch size of 32. The training was conducted for 500 epochs. The final output of the model is a denoised PET image, which aims to approximate the ground truth full count image. The model was trained by MSE objective function.

\section{RESULTS}
We conducted experiments using the publicly available Siemens Biograph Vision Quadra dataset provided by the Ultra-low Count PET Imaging Challenge~\cite{xue2021crossscanner}. This dataset comprises 387 total-body 18F-FDG PET scans. All data were ac-quired in list mode, enabling retrospective rebinding to simulate various acquisition durations. Simulated low-count PET images, representing reduced dose scenarios with specific dose reduction factor, were reconstructed by resampling counts from a shortened time window centered within the original acquisition period. The resulting images include reconstructions at 1/100, 1/20, 1/10, 1/4, and 1/2 of the full count, as well as the full count image. These varying count levels provide a robust foundation for training and evaluating count-aware denoising models.
We used various count level images in the test images to specifically synthesize the full count image with output text of full count level. The visual example of synthetic denoising full count image from various count levels shown in Figure 2. The proposed model can successfully denoise various count levels images to full count PET images. For quantitative evaluation, we denoised images at all count levels to generate synthetic full-count images, then compared the SSIM and PSNR between the original low-count images and the corresponding synthetic full-count images, using the real full-dose image as the reference. The quantitative results which are shown in Figure 3 indicate that denoised images from different count levels are closer to the full count image than the original low count image. We also compared model ability for synthetic full count image from 1/100 count image with U-Net and CycleGAN model, for just testing the ability of the flexible framework, only the baseline methods were compared. Note that the U-Net and CycleGAN model only trained with 1/100 count model and regard full count image as ground truth. As shown in Figure 4, regarding the visual results, the proposed method is closer to the full count, and the PSNR and SSIM of the proposed method is higher than the U-Net and CycleGAN models.

\begin{figure} [htbp]
\begin{center}
\begin{tabular}{c} 
\includegraphics[width=16cm]{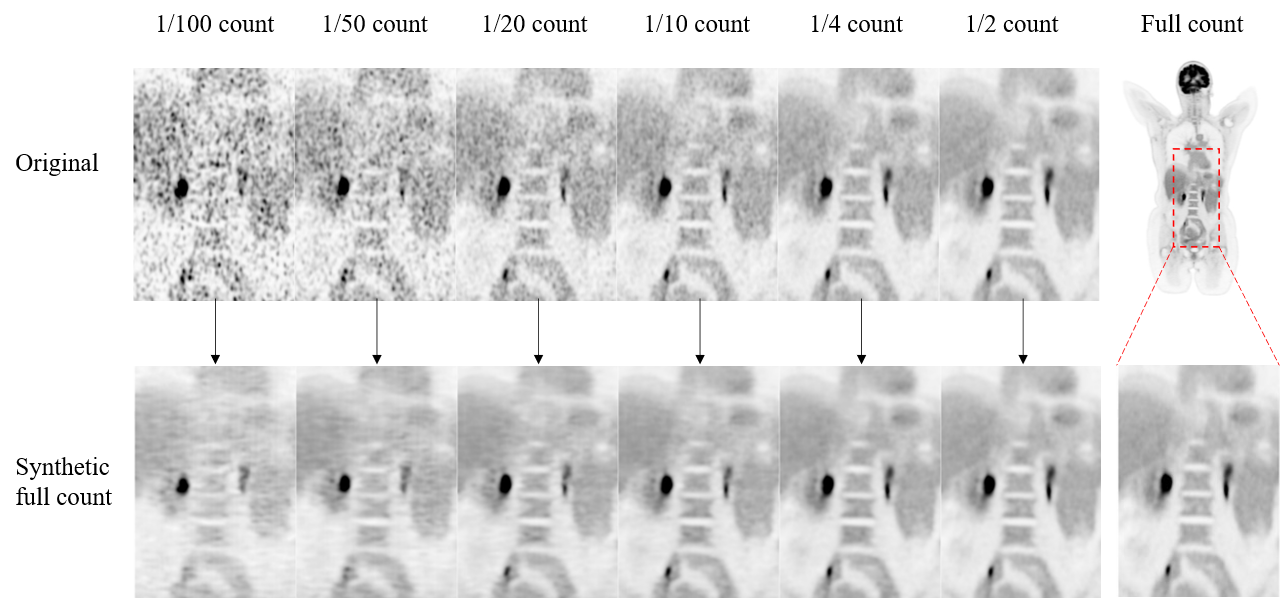}
\end{tabular}
\end{center}
\caption[Fig2]
{The visual example of synthetizing full count image from various count levels. The first row is the original count level images, the second row is the synthetic full count image from the corresponding count level with the proposed model. }
\end{figure}

\begin{figure} [htbp]
\begin{center}
\begin{tabular}{c} 
\includegraphics[width=12cm]{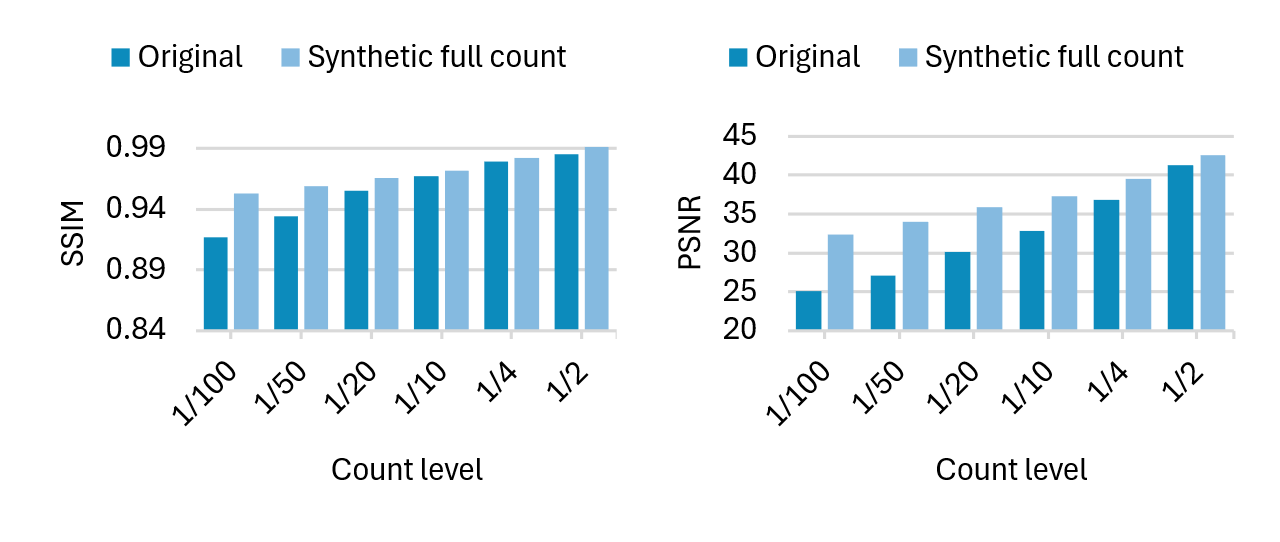}
\end{tabular}
\end{center}
\caption[Fig3]
{The SSIM and PSNR of original various count vs. full count and synthetic full count from various count level vs. full count.}
\end{figure}

\begin{figure} [htbp]
\begin{center}
\begin{tabular}{c} 
\includegraphics[width=16cm]{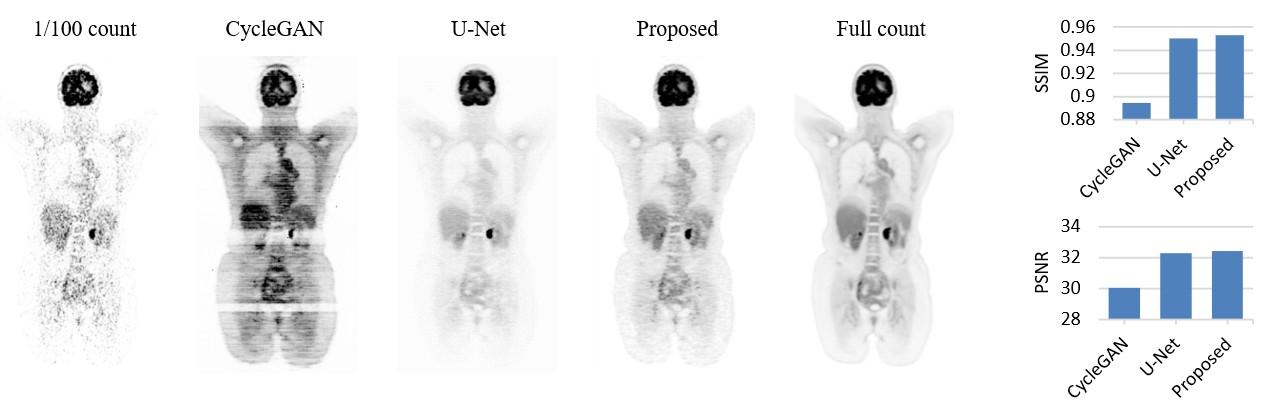}
\end{tabular}
\end{center}
\caption[Fig4]
{Visual and quantification comparison between the original 1/100 count image, denoised using CycleGAN, U-Net and the proposed method (all from the 1/100 count image), with the full count image as reference.}
\end{figure}

\section{CONCLUSIONS}
This approach provides flexible and controllable denoising across a wide range of input–output count scenarios, overcoming the limitations of fixed-pair training in conventional models. By conditioning the denoising process on count-level embeddings, the model adaptively reconstructs PET images that reflect the characteristics of the desired count level. Furthermore, this design supports count-level generalization, enabling synthesis of PET images at arbitrary higher count levels from a single low-count input. Such flexibility is particularly valuable in clinical settings where acquisition protocols vary and minimizing radiation exposure is critical.

In this study, we introduce a novel text-controlled PET denoising model that integrates dual text embeddings into both the encoder and decoder paths of a U-Net-based architecture. Leveraging semantic information from count-level descriptions via a pretrained CLIP text encoder, the model can synthesize PET images across diverse count levels. Unlike traditional approaches constrained by fixed input–output pairs, our method enables dynamic count-level conditioning, allowing low-count PET images to be flexibly enhanced to approximate higher-count counterparts. This design improves generalizability across varying count levels and supports applications in count-aware image synthesis and low-count PET enhancement for diagnostic purposes.

To assess the effectiveness of the proposed model, and given the absence of an existing baseline with similar functionality, we compared it with U-Net and CycleGAN under the same task of denoising 1/100-count images to full-dose equivalents. Experimental results demonstrate that our model achieves superior performance in terms of SSIM and PSNR across multiple count levels, validating its adaptability and robustness. This work introduces new possibilities for count-controllable PET reconstruction, which is particularly valuable in clinical and research settings where balancing radiation exposure and image quality is critical. However, the current study is limited by the availability of paired input–output data. In future work, with access to sufficient raw list-mode PET data, arbitrary dose-level PET images could be simulated, further enhancing the flexibility and applicability of the proposed approach.

\section*{STATEMENT} 
This work is original and in its present form has not been submitted elsewhere in any form. The authors would like to extend their sincere thanks to Xue Song, Kuangyu Shi, and Axel Rominger from the Department of Nuclear Medicine, University of Bern, as well as Hanzhong Wang, Rui Guo, and Biao Li from Ruijin Hospital, Shanghai Jiao Tong University, for generously providing the data used in this study. All authors affirm that they have no conflicts of interest financial or personal that could have influenced the outcomes or interpretations presented in this work. The data employed in this research were sourced from the Department of Nuclear Medicine, University of Bern, and the School of Medicine, Ruijin Hospital. While the contributing investigators were involved in the design and development of the dataset and/or data provision, they did not participate in the analysis or manuscript preparation. A comprehensive list of investigators is available at: “https://ultra-low-dose-pet.grand-challenge.org/Description/”.

\bibliography{main} 
\bibliographystyle{spiebib} 

\end{document}